\def\year{2018}\relax
\documentclass[letterpaper]{article} 
\usepackage{aaai18}  
\usepackage{times}  
\usepackage{helvet}  
\usepackage{courier}  
\usepackage{url}  
\usepackage{graphicx}  
\usepackage[ruled, linesnumbered]{algorithm2e}
\usepackage{amsfonts}
\usepackage{amsmath}
\usepackage[title]{appendix}
\usepackage{booktabs}
\usepackage{multirow}
\usepackage{subcaption}
\usepackage{tikz}

\usetikzlibrary{shapes}
\usetikzlibrary{fit}
\usetikzlibrary{chains}
\usetikzlibrary{arrows}

\tikzstyle{latent} = [circle,fill=white,draw=black,inner sep=1pt,
minimum size=20pt, font=\fontsize{10}{10}\selectfont, node distance=1]
\tikzstyle{obs} = [latent,fill=gray!25]
\tikzstyle{const} = [rectangle, inner sep=0pt, node distance=1]
\tikzstyle{factor} = [rectangle, fill=black,minimum size=5pt, inner
sep=0pt, node distance=0.4]
\tikzstyle{det} = [latent, diamond]

\tikzstyle{plate} = [draw, rectangle, rounded corners, fit=#1]
\tikzstyle{wrap} = [inner sep=0pt, fit=#1]
\tikzstyle{gate} = [draw, rectangle, dashed, fit=#1]

\tikzstyle{caption} = [font=\footnotesize, node distance=0] %
\tikzstyle{plate caption} = [caption, node distance=0, inner sep=0pt,
below left=5pt and 0pt of #1.south east] %
\tikzstyle{factor caption} = [caption] %
\tikzstyle{every label} += [caption] %

\tikzset{>={triangle 45}}

\newcommand{\edge}[3][]{ %
  \foreach \x in {#2} { %
    \foreach \y in {#3} { %
      \path (\x) edge [->,#1] (\y) ;%
    } ;
  } ;
}

\begin{document}
%

\title{Generating Sentences Using a Dynamic Canvas}
\author{Harshil Shah \\ University College London
\And Bowen Zheng \\ University College London
\And David Barber \\ University College London \\ \& \\ Alan Turing Institute
}

\maketitle

\begin{abstract}
We introduce the \textit{\textbf{A}ttentive \textbf{U}nsupervised \textbf{T}ext (W)\textbf{r}iter} (AUTR), which is a word level generative model for natural language. It uses a recurrent neural network with a dynamic attention and canvas memory mechanism to iteratively construct sentences. By viewing the state of the memory at intermediate stages and where the model is placing its attention, we gain insight into how it constructs sentences. We demonstrate that AUTR learns a meaningful latent representation for each sentence, and achieves competitive log-likelihood lower bounds whilst being computationally efficient. It is effective at generating and reconstructing sentences, as well as imputing missing words.
\end{abstract}

\section{Introduction}

Latent variable models have recently enjoyed significant success when modelling images \cite{GDGRW2015,REMBJH2016,GKATVVC2017}, as well as sequential data such as handwriting and speech \cite{BO2015,CKDGCB2015}. They specify a conditional distribution of observed data, given a set of hidden (latent) variables. The stochastic gradient variational Bayes (SGVB) algorithm \cite{KW2014,RMW2014} has made (approximate) maximum likelihood learning possible on a large scale in models where the true posterior distribution of the latent variables is not tractable. Deep neural networks can be used to parametrise the generative and variational distributions, allowing for extremely flexible and powerful model classes.

There has been somewhat less exploration into deep generative models for natural language. \citeauthor{G2014} \shortcite{G2014} uses a stacked RNN architecture at the character level to generate sentences with long range dependencies. The model sequentially emits characters based on the previously generated ones, however it does not map each sentence to a single latent representation. This means that even though the generated sentences are syntactically coherent and may show local semantic consistency, the model does not encourage the sentences to have long range semantic consistency. Additionally, the model cannot generate sentences conditioned on meaning, style, etc. \citeauthor{BVVDJB2016} \shortcite{BVVDJB2016} use a word level latent variable model with an RNN and train it using SGVB (we refer to this as Gen-RNN). Samples from the prior produce well-formed, coherent sentences, and the model is effective at imputing missing words. However, the authors find that the KL divergence term of the log-likelihood lower bound reduces to 0, which implies that the model ignores the latent representation and collapses to a standard RNN language model, similar to that of \citeauthor{G2014} \shortcite{G2014}. The authors use word dropout to alleviate this problem, and show that Gen-RNN generates sentences with more varied vocabulary and is better at imputing missing words than the RNN language model. \citeauthor{SSB2017} \shortcite{SSB2017} and \citeauthor{YHSB2017} \shortcite{YHSB2017} make use of convolutional layers, which appear to encourage their models to more strongly rely on the latent representation without using word dropout.

In the context of computer vision, DRAW \cite{GDGRW2015} showed that using an attention mechanism to `paint' locally on a canvas produced images of remarkable quality. A natural question therefore is whether a similar approach could work well for natural language. To this end we introduce the \textit{\textbf{A}ttentive \textbf{U}nsupervised \textbf{T}ext (W)\textbf{r}iter} (AUTR), which is a word level generative model for text; AUTR uses an RNN with a dynamic attention mechanism to iteratively update a canvas (analogous to an external memory \cite{GHSWMRAL2017}).

Using an attention mechanism in this way can be very powerful---it allows the model to use the RNN to focus on local parts of the sentence at each time step whilst relying on the latent representation to encode global sentence features. By viewing the canvas at intermediate stages, and where the RNN is placing its attention, we gain insight into how the model constructs a sentence. Additionally, we verify that AUTR attains competitive lower bounds on the log-likelihood whilst being computationally efficient. As well as learning a meaningful latent representation for each sentence, the model generates coherent sentences and successfully imputes missing words. A generative model which is able to, in some sense, `understand' natural language (which we believe AUTR shows signs of doing) should facilitate much better performance when used as a module for downstream tasks such as translation and question answering.

The remainder of this paper is structured as follows: in section \ref{sec:model} we define the AUTR architecture and generative process, in section \ref{sec:related_work} we review related work, in section \ref{sec:experiments} we provide experimental results on the Book Corpus dataset along with examples of generated sentences, and in section \ref{sec:conclusion} we make concluding remarks. Appendix \ref{sec:variational_inference} provides a review of the SGVB algorithm.

\section{Model} \label{sec:model}

AUTR is a word level generative recurrent neural network (RNN) which iteratively updates a canvas that parametrises the probability distribution over the sentence's text. Using \( L \) to denote the number of words in the sentence and \( E \) to denote the word embedding size, the canvas \( \mathbf{C} \in \mathbb{R}^{L \times E} \) is a 2 dimensional array with \( L \) `slots', each of which represents the model's estimation of the word embedding for that position in the sentence.

We use \( T \) to denote the number of time steps in the RNN---at each time step an attention mechanism selects the canvas' slots to be updated; \( \mathbf{C}^{t} \) denotes the state of the canvas at time step \( t \). Note that \( T \) is a hyper-parameter of the model.

\begin{figure}[t]
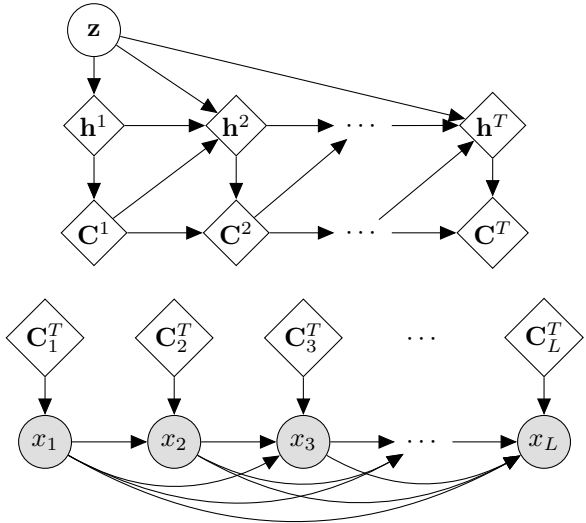

    \centering
    \tikz[scale=0.5]{%
        \node[det] (C^T) {$\mathbf{C}^{T}$};
        \node[det, above=0.5 of C^T] (h^T) {$\mathbf{h}^{T}$};
        \edge {h^T} {C^T};
        \node[const, left=of C^T] (C^dots) {$\cdots$};
        \edge[shorten <=4pt] {C^dots} {C^T};
        \edge[shorten <=4pt] {C^dots} {h^T};
        \node[const, above=0.5 of C^dots, yshift=21.5pt] (h^dots) {$\cdots$};
        \edge[shorten <=4pt] {h^dots} {h^T};
        \node[det, left=of C^dots] (C^2) {$\mathbf{C}^{2}$};
        \edge[shorten >=4pt] {C^2} {C^dots};
        \edge[shorten >=4pt] {C^2} {h^dots};
        \node[det, above=0.5 of C^2, yshift=2pt] (h^2) {$\mathbf{h}^{2}$};
        \edge[shorten >=4pt] {h^2} {h^dots};
        \edge {h^2} {C^2};
        \node[det, left=of C^2] (C^1) {$\mathbf{C}^{1}$};
        \edge {C^1} {C^2};
        \edge {C^1} {h^2};
        \node[det, above=0.5 of C^1, yshift=2pt] (h^1) {$\mathbf{h}^{1}$};
        \edge {h^1} {h^2};
        \edge {h^1} {C^1};
        \node[latent, above=0.5 of h^1] (z) {$\mathbf{z}$};
        \edge {z} {h^1};
        \edge {z} {h^2};
        \edge {z} {h^T};
    } \\[10pt]
    \tikz[scale=0.5]{%
        \node[obs] (x_L) {$x_{L}$};
        \node[det, above=0.5 of x_L] (C_L) {$\mathbf{C}^{T}_{L}$};
        \edge {C_L} {x_L};
        \node[const, left=of x_L] (x_dots) {$\cdots$};
        \edge[shorten <=4pt] {x_dots} {x_L};
        \node[const, above=0.5 of x_dots, yshift=20pt] (C_dots) {$\cdots$};
        \node[obs, left=of x_dots] (x_3) {$x_{3}$};
        \edge[shorten >=4pt] {x_3} {x_dots};
        \edge[bend right=30] {x_3} {x_L};
        \node[det, above=0.5 of x_3] (C_3) {$\mathbf{C}^{T}_{3}$};
        \edge {C_3} {x_3};
        \node[obs, left=of x_3] (x_2) {$x_{2}$};
        \edge{x_2} {x_3};
        \edge[bend right=30, shorten >=4pt] {x_2} {x_dots};
        \edge[bend right=30] {x_2} {x_L};
        \node[det, above=0.5 of x_2] (C_2) {$\mathbf{C}^{T}_{2}$};
        \edge {C_2} {x_2};
        \node[obs, left=of x_2] (x_1) {$x_{1}$};
        \edge{x_1} {x_2};
        \edge[bend right=30, shorten >=4pt] {x_1} {x_dots};
        \edge[bend right=30] {x_1} {x_3};
        \edge[bend right=30] {x_1} {x_L};
        \node[det, above=0.5 of x_1] (C_1) {$\mathbf{C}^{T}_{1}$};
        \edge {C_1} {x_1};
    }
    \caption{The AUTR generative model. The shaded nodes signify the observed words and non-shaded nodes signify latent random variables; a rhombus node is a deterministic function of its incoming variables. Here \( \mathbf{z} \) denotes the latent sentence representation; \( \mathbf{C}^{t} \ \forall \ t \in \{1, \hdots, T\} \) are the states of the canvas; \( \mathbf{C}^{T}_{l} \ \forall \ l \in \{1, \hdots, L\} \) are the slots of the final canvas corresponding to each word; \( x_{l} \ \forall \ l \in \{1, \hdots, L\} \) are the observed words.}
    \label{fig:generative_model}
\end{figure}

Figure \ref{fig:generative_model} shows the graphical model for AUTR and figure \ref{fig:prior_canvases} shows examples of how AUTR iteratively constructs sentences. Like other latent variable models \cite{BVVDJB2016,SSB2017,YHSB2017}, AUTR uses a hidden representation \( \mathbf{z} \) to encode each sentence, and can construct new sentences by sampling \( \mathbf{z} \) from a prior distribution \( p(\mathbf{z}) \) and passing this sample through the RNN. A summary of the generative process is given in algorithm \ref{alg:generative_process} and full details follow in section \ref{sec:generative_process}.

\begin{algorithm}[h]
    
    \SetAlgoLined
    
    Sample a latent vector \( \mathbf{z} \) from a \( \mathcal{N}(\mathbf{0}, \mathbf{I}) \) distribution.
    
    Initialise both the hidden state and canvas as zeros, \textit{i.e.} \( \mathbf{h}^{0} = \mathbf{0} \) and \( \mathbf{C}^{0} = \mathbf{0} \)
    
    \For{\( t = 1, \hdots, T \)}{%
        Compute the hidden state: \( \mathbf{h}^{t} = f(\mathbf{z}, \mathbf{h}^{t-1}, \mathbf{C}^{t-1}) \)
        
        Compute the gate: \( \mathbf{g}^{t}_{l} = \frac{\exp \left[ (\mathbf{W}_{g} \cdot \mathbf{h}^{t})_{l} \right] \left( 1 - \sum_{t'=1}^{t-1} \mathbf{g}^{t'}_{l} \right)}{\sum_{k=1}^{L} \exp \left[ (\mathbf{W}_{g} \cdot \mathbf{h}^{t})_{k} \right] \left( 1 - \sum_{t'=1}^{t-1} \mathbf{g}^{t'}_{k} \right)} \left( 1 - \sum_{t'=1}^{t-1} \mathbf{g}^{t'}_{l} \right) \) for \( l = 1, \hdots, L \)
        
        Update the canvas: \( \mathbf{C}^{t} = (\mathbf{1} - \mathbf{g}^{t}) \odot \mathbf{C}^{t-1} + \mathbf{g}^{t} \odot (\mathbf{W}_{u} \cdot \mathbf{h}^{t}) \)
    }
    
    Sample the sentence according to the distribution: \( \mathbf{x} \sim p(\mathbf{x}|\mathbf{z}, \mathbf{C}^{T}) = p(x_{1}|\mathbf{z}, \mathbf{C}^{T}_{1}) \prod_{l=2}^{L} p(x_{l}|\mathbf{z},\mathbf{C}^{T}_{l}, x_{1:l-1}) \)
    
    \caption{AUTR generative process}
    \label{alg:generative_process}
    
\end{algorithm}

\subsection{Generative process} \label{sec:generative_process}

We first sample the latent representation \( \mathbf{z} \) from a \( \mathcal{N}(\mathbf{0}, \mathbf{I}) \) distribution. Each RNN hidden state is then computed as a function of this latent representation, as well as the previous RNN hidden state and the canvas so far: \( \mathbf{h}^{t} = f(\mathbf{z}, \mathbf{h}^{t-1}, \mathbf{C}^{t-1}) \). In our experiments, we use the LSTM for \( f(\cdot) \) \cite{HS1997}. Allowing the RNN hidden state to see what has been written to the canvas so far allows the model to maintain the sentence's long range semantic coherence because the hidden state can anticipate the words that will be written at the end of the sentence and adjust the beginning accordingly, and vice versa.

Each hidden state \( \mathbf{h}^{t} \) is then used to determine where to write (or more specifically, how `strongly' to write to each of the canvas' slots). We denote the gate as \( \mathbf{g}^{t}_{l} \in [0, 1] \) for \( l = 1, \hdots, L \). 

\subsubsection{Attention mechanism} \label{sec:attention}

For the gate (or attention), we use a modified softmax attention mechanism. A standard softmax mechanism \cite{LPM2015} would be: \begin{align}
    \mathbf{g}^{t}_{l} = \frac{\exp \left[ (\mathbf{W}_{g} \cdot \mathbf{h}^{t})_{l} \right]}{\sum_{k=1}^{L} \exp \left[ (\mathbf{W}_{g} \cdot \mathbf{h}^{t})_{k} \right]}
\end{align}

This would ensure that (at each time step) the attention for each slot is between 0 and 1 and the total attention across all slots is 1. To encourage the model to write to those slots where it hasn't yet written, we multiply the elements of the softmax by \( (1 - \sum_{t'=1}^{t-1} \mathbf{g}^{t'}_{l}) \). To ensure that the cumulative attention, over time, applied to any of the slots is no greater than 1, we multiply the softmax itself by \( (1 - \sum_{t'=1}^{t-1} \mathbf{g}^{t'}_{l}) \). This results in the following modified attention mechanism: \begin{align}
    \mathbf{g}^{t}_{l} = \frac{\exp \left[ (\mathbf{W}_{g} \cdot \mathbf{h}^{t})_{l} \right] s_{l}}{\sum_{k=1}^{L} \left[ \exp \left[ (\mathbf{W}_{g} \cdot \mathbf{h}^{t})_{k} \right] s_{k} \right]} \cdot s_{l}
\end{align}

\noindent where \( s_{l} = \left( 1 - \sum_{t'=1}^{t-1} \mathbf{g}^{t'}_{l} \right) \). In our experiments, we found that this modification performed favourably compared to the standard softmax mechanism. Note that, once a slot has been written to with a cumulative attention of 1 (\textit{i.e.} \( \sum_{t'=1}^{t-1} \mathbf{g}^{t'}_{l} = 1 \)), it cannot be updated further.

One of the key computational advantages of AUTR is that it works well with \( T < L \), because it writes to multiple slots at each RNN time step. This is shown to work well empirically in section \ref{sec:results}.

\subsubsection{Updating the canvas}

The content to be written to the canvas is a linear function of the hidden state at that time step: \( \mathbf{U}^{t} = \mathbf{W}_{u} \cdot \mathbf{h}^{t} \).

Using \( \odot \) to denote element-wise multiplication, the canvas is updated as: \( \mathbf{C}^{t} = (\mathbf{1} - \mathbf{g}^{t}) \odot \mathbf{C}^{t-1} + \mathbf{g}^{t} \odot \mathbf{U}^{t} \).

\( \mathbf{g}^{t}_{l} = 0 \) means that the \( l^{\mathrm{th}} \) slot from the previous time step carries over exactly to the current time step (\textit{i.e.} no updating takes place), whereas \( \mathbf{g}^{t}_{l} = 1 \) means that the previous values of the \( l^{\mathrm{th}} \) slot are completely forgotten and new values are entered in their place.

We tested a fixed mechanism instead of one with attention to update the canvas, but found that the RNN didn't use the \( T \) computational steps available - at the final time step it simply overwrote everything it had previously written.

\subsubsection{Text generation}

Conditioned on the final canvas, we considered first a model that generated each word independently. However, this was ineffective in our experiments since local consistency between words was lost. Therefore, we sample the sentence's text from a Markov model where the sampled word at position \( l \) depends on the \( l^{\textrm{th}} \) slot of the final canvas \( \mathbf{C}^{T}_{l} \) and on all of the \( l - 1 \) words that have been sampled so far. We first compute the `context' \( \tilde{\mathbf{x}}_{l} \), which is a weighted average of the previously generated words; this is then used to modify the word probabilities that would have been assigned by the canvas alone. Specifically: \begin{align}
    \tilde{\mathbf{x}}_{l} = \sum_{l'=1}^{l-1} w^{l}_{l'}(\mathbf{z}) \mathbf{e}(x_{l'})
\end{align}
where \( \mathbf{e}(x) \) is the embedding of word \( x \) and: \begin{align}
    w^{l}_{l'}(\mathbf{z}) = \frac{\exp[(\mathbf{W}_{l} \cdot \mathbf{z})_{l'}]}{\sum_{l'=1}^{l-1} \exp[(\mathbf{W}_{l} \cdot \mathbf{z})_{l'}]} 
\end{align}

Therefore \( \sum_{l'=1}^{l-1} w^{l}_{l'}(\mathbf{z}) = 1 \). Then, denoting \( \mathbf{b}_{l} = \mathbf{C}^{T}_{l} + \mathbf{W}_{x} \cdot \tilde{\mathbf{x}}_{l} + \mathbf{W}_{z} \cdot \mathbf{z} \):
 \begin{align}
    p(x_{l}=a|\mathbf{z},x_{1:l-1},\mathbf{C}^{T}_{l}) = \frac{\exp[\mathbf{e}(a) \cdot \mathbf{b}_{l}]}{\sum_{v \in \mathcal{V}} \exp[\mathbf{e}(v) \cdot \mathbf{b}_{l}]} \label{eq:dist_x}
\end{align}

\noindent where \( \mathcal{V} \) is the entire vocabulary. 

\subsection{Inference} \label{sec:model_sgvb}

Due to the intractability of the true posterior \( p(\mathbf{z}|\mathbf{x}) \), we perform (approximate) maximum likelihood estimation using SGVB, as described in appendix \ref{sec:variational_inference}. The variational distribution is Gaussian: \( q_{\phi}(\mathbf{z}|\mathbf{x}) = \mathcal{N}(\pmb{\mu}_{\phi}(\mathbf{x}), \mathrm{diag}(\pmb{\sigma}_{\phi}^{2}(\mathbf{x}))) \); its mean and variance are parametrised by an RNN which takes as input each word embedding in order, one at a time \cite{BVVDJB2016}. For the hidden states, we use the LSTM and the final hidden state is passed through a feedforward network with two output layers to produce the mean and variance of the variational distribution.

\section{Related Work} \label{sec:related_work}

The use of stochastic gradient variational Bayes (SGVB) \cite{KW2014,RMW2014} to train latent variable generative models is widespread for images \cite{GDGRW2015,REMBJH2016,GKATVVC2017}.

Whilst latent variable generative models for natural language have been less common, they have recently increased in popularity. \citeauthor{BVVDJB2016} \shortcite{BVVDJB2016} use an LSTM for the generative model (Gen-RNN), which outputs a single word at each time step. This is in contrast to AUTR, which updates the distribution for every word in the sentence at each RNN time step, using an attention mechanism.

More recently, \citeauthor{MGB2017} \shortcite{MGB2017} use an RNN based generative model similar to Gen-RNN in order to perform topic allocation to documents. \citeauthor{YHSB2017} \shortcite{YHSB2017} use dilated convolutions to replace the recurrent structure in Gen-RNN in order to control the contextual capacity, leading to better performance.

Our canvas based model with its dynamic attention mechanism is largely inspired by DRAW \cite{GDGRW2015}, which iteratively updates the canvas that parametrises the final distribution over the observed image. One of the primary differences between DRAW and AUTR is that conditioned on the final canvas DRAW treats each pixel independently---whilst this may not be too constraining in the image domain, this was ineffective in our natural language experiments where the sampled word embedding may differ significantly from the entry in that slot of the canvas. Therefore AUTR conditions on all previously generated words in the sentence when sampling the next word.

\section{Experiments} \label{sec:experiments}

\begin{figure*}[t!]
    \centering
    \includegraphics[width=\textwidth]{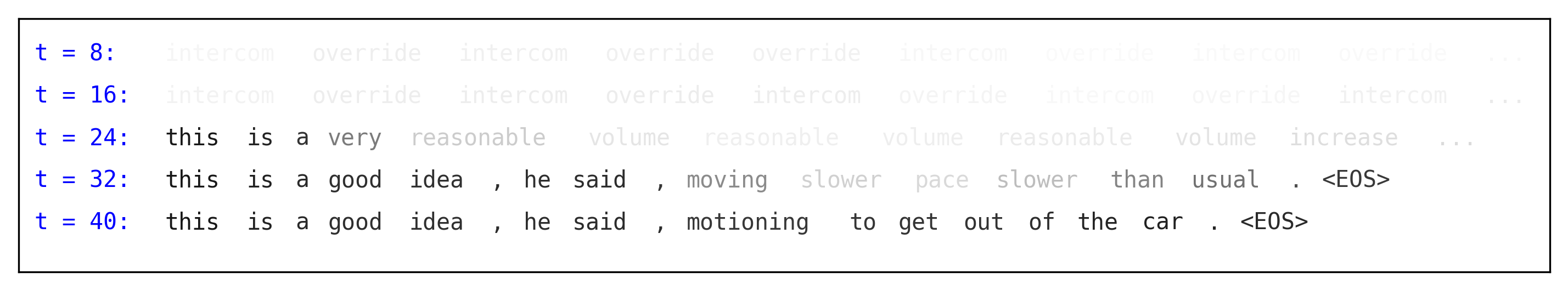}
    \includegraphics[width=\textwidth]{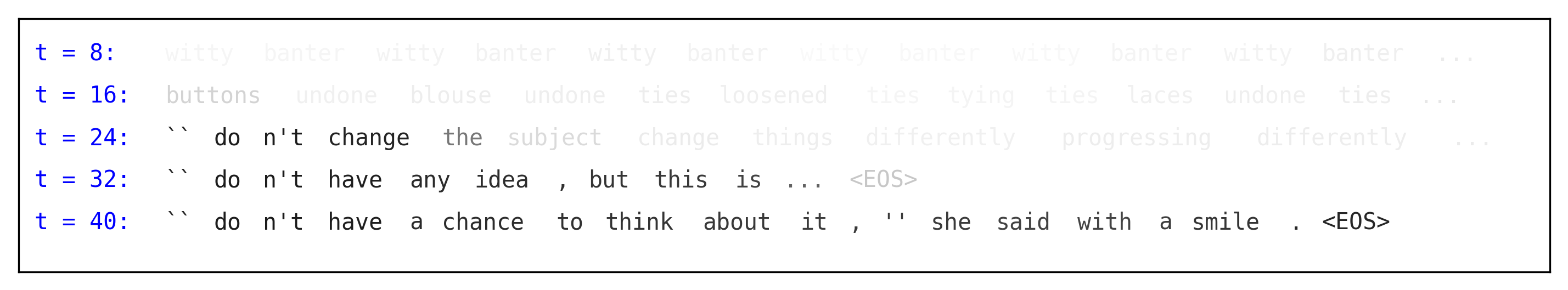}
    \caption{Visualising the sequential construction of sentences generated from the learned model. We sample \( \mathbf{z} \) from its prior, \textit{i.e.} \( \mathbf{z} \sim \mathcal{N}(\mathbf{0}, \mathbf{I}) \) and pass the sample through the RNN, visualising the canvas at several points along the way. The darkness indicates the cumulative attention that has been placed on that slot so far.}
    \label{fig:prior_canvases}
\end{figure*}

We train our model on the Book Corpus dataset \cite{ZKZSUTF2015}, which is composed of sentences from 11,038 unpublished books. We report results on language modelling tasks, comparing against Gen-RNN.

\subsection{Preprocessing}

We restrict the vocabulary size to 20,000 words and we use sentences with a maximum length of 40 words. Of the 53M sentences that meet these criteria, we use 90\% for training, and 10\% for testing.

\subsection{Model architectures}

\subsubsection{Generative model}

For both AUTR and Gen-RNN, we use a 50 dimensional latent representation \( \mathbf{z} \) and the RNN hidden states have 500 units each. To compute the hidden states, we use the LSTM \cite{HS1997}.

Gen-RNN requires \( T = L = 40 \), because it outputs one word at each RNN time step. However, as explained in section \ref{sec:attention}, one of the key advantages of AUTR is that it works well with \( T < L = 40 \). Therefore, for AUTR, we compare results with \( T \in \{30, 40\} \).

\subsubsection{Variational distribution}

As per section \ref{sec:model_sgvb}, we optimise both AUTR and Gen-RNN using SGVB. In both models, we use the LSTM architecture introduced by \citeauthor{BVVDJB2016} \shortcite{BVVDJB2016} to parametrise the mean and variance of the Gaussian variational distribution; the hidden states have 500 units each.

\subsubsection{Parameters and speed} For both models, we use 300 dimensional word embeddings, which are learned jointly with the generative and variational parameters. AUTR and Gen-RNN have 10.9M and 10.3M parameters respectively. They take, on average, 0.19 and 0.17 seconds per training iteration and 0.06 and 0.05 seconds for sentence generation respectively.\footnotemark \footnotetext{These values are for AUTR with \( T = 40 \).}

\subsection{Training process}

We optimise the ELBO, shown in equation (\ref{eq:elbo}), using stochastic gradient ascent. We train both models for 1,000,000 iterations, using Adam \cite{KB2015} with an initial learning rate of \( 10^{-4} \) and  mini-batches of size 200. To ensure training is fast, we use only a single sample \( \mathbf{z} \) per data point from the variational distribution at each iteration. We implement both models in Python, using the Theano \cite{T2016} and Lasagne \cite{DSROS2015} libraries.

\textbf{KL divergence annealing} \ The ELBO can be expressed as: \( \mathcal{L}(\mathbf{x}) = \mathbb{E}_{q_{\phi}(\mathbf{z}|\mathbf{x})} \left[ \log p(\mathbf{x}|\mathbf{z}) \right] - D_{\mathrm{KL}} \left[ q_{\phi}(\mathbf{z}|\mathbf{x}) \ || \ p(\mathbf{z}) \right] \). We multiply the KL divergence term by a constant weight, which we linearly anneal from 0 to 1 over the first 20,000 iterations of training \cite{BVVDJB2016,SRMSW2016}.

\textbf{Word dropout} \ To encourage Gen-RNN to make better use of the latent representation, \citeauthor{BVVDJB2016} \shortcite{BVVDJB2016} randomly drop out a proportion of the words when training the generative RNN - without this, they find that their model collapses to a simple RNN language model which ignores the latent representation. Following this, when training Gen-RNN, we randomly drop out 30\% of the words. However, to show that (unlike Gen-RNN) AUTR does not need the dropout mechanism to avoid the KL divergence term \( D_{\mathrm{KL}} \left[ q_{\phi}(\mathbf{z}|\mathbf{x}) \ || \ p(\mathbf{z}) \right] \) from collapsing to 0, we train it both with 30\% dropout, and without any dropout.

\subsection{Results} \label{sec:results}

We report test set results on the Book Corpus dataset in table \ref{tab:results}. We evaluate the ELBO on the test set by drawing 1,000 samples of the latent vector \( \mathbf{z} \) per data point. We see that AUTR, both with \( T = 30 \) and \( T = 40 \), trained with or without dropout, achieves a higher ELBO and lower perplexity than Gen-RNN. Importantly, AUTR (trained with and without dropout) relies more heavily on the latent representation than Gen-RNN, as is shown by the larger contribution to the ELBO from the KL divergence term. Note that if a model isn't taking advantage of the latent vector \( \mathbf{z} \), the loss function drives it to set \( q(\mathbf{z}|\mathbf{x}) \) equal to the prior on \( \mathbf{z} \) (disregarding \( \mathbf{x} \)), which yields a KL divergence of zero.

\begin{table}[h]
    \centering
    \begin{tabular}{ c c c c c }
        \toprule
        \multicolumn{2}{c}{Model} & ELBO & KL & PPL \\[2pt]
        \midrule
        \multicolumn{2}{c}{Gen-RNN (30\% dropout)} & -52.3 & 7.1 & 41.9 \\[2pt]
        \multirow{2}{*}{AUTR (no dropout)} & T = 30 & -50.7 & 8.0 & 37.4 \\[2pt]
         & T = 40 & -50.7 & 7.8 & 37.4 \\[2pt]
        \multirow{2}{*}{AUTR (30\% dropout)} & T = 30 & -51.6 & 14.0 & 39.9 \\[2pt]
         & T = 40 & -51.5 & 13.8 & 39.6 \\[2pt]
        \bottomrule
    \end{tabular}
    \caption{Test set results on the Book Corpus dataset. We report the ELBO, the contribution to the ELBO from the KL divergence term \( (D_{\mathrm{KL}} \left[ q_{\phi}(\mathbf{z}|\mathbf{x}) \ || \ p(\mathbf{z}) \right]) \), and the perplexity (PPL) on the test set. For the ELBO, higher is better, and for the perplexity, lower is better.}
    \label{tab:results}
\end{table}

\subsection{Observing the generation process} \label{sec:observe_gen_process}

\begin{table*}[t!]
    \centering
    \begin{subtable}[b]{0.55\textwidth}
        \centering
        \begin{tabular}{ @{} l @{} }
            \toprule
            AUTR \\
            \toprule
            ``do you have any idea how much i love you when i'm with you?'' \\
            \midrule
            ``if he didn't want to kill me,'' he said , but he was trying to keep \\
            distance. \\
            \midrule
            hundreds of thousands of stars rose above, reflecting the sky above \\
            the horizon. \\
            \midrule
            ``that sounds like a good idea,'' he said, as his voice trailed off. \\
            \bottomrule
        \end{tabular}
    \end{subtable}%
    \begin{subtable}[b]{0.45\textwidth}
        \centering
        \begin{tabular}{ @{} l @{} }
            \toprule
            Gen-RNN \\
            \toprule
            but i didn't want to think of any other way to get it. \\
            \midrule
            when i reach the top of the stairs , i feel of sight \\
            of my back into the doors open the door swings . \\
            \midrule
            i had no idea what i was going to do, but i was wrong. \\
            \midrule
            ``you're going to look at least one of course, but i \\
            have been in the most of course,'' he said. \\
            \bottomrule
        \end{tabular}
    \end{subtable}
    \caption{Sentences sampled from the prior: \( \mathbf{z} \) is drawn from \( \mathcal{N}(\mathbf{0}, \mathbf{I}) \) and passed through the generative model \( p_{\theta}(\mathbf{x}|\mathbf{z}) \) to produce a sentence \( \mathbf{x} \).}
    \label{tab:samples:prior}
\end{table*}

Conditioned on a sampled \( \mathbf{z} \), we would like to know the most likely sentence, \textit{i.e.} \( \arg \max_{\mathbf{x}} \log p_{\theta}(\mathbf{x}|\mathbf{z}) \). However, because each word depends on all of the words generated before it, this optimisation has a computationally intractable memory requirement. We therefore perform this maximisation approximately by considering only the \( K \) `best' trajectories for each position in the sentence - this is known as beam search with beam size \( K \) \cite{WR2016}.

\begin{figure}[h]
    \centering
    \includegraphics[width=0.9\columnwidth]{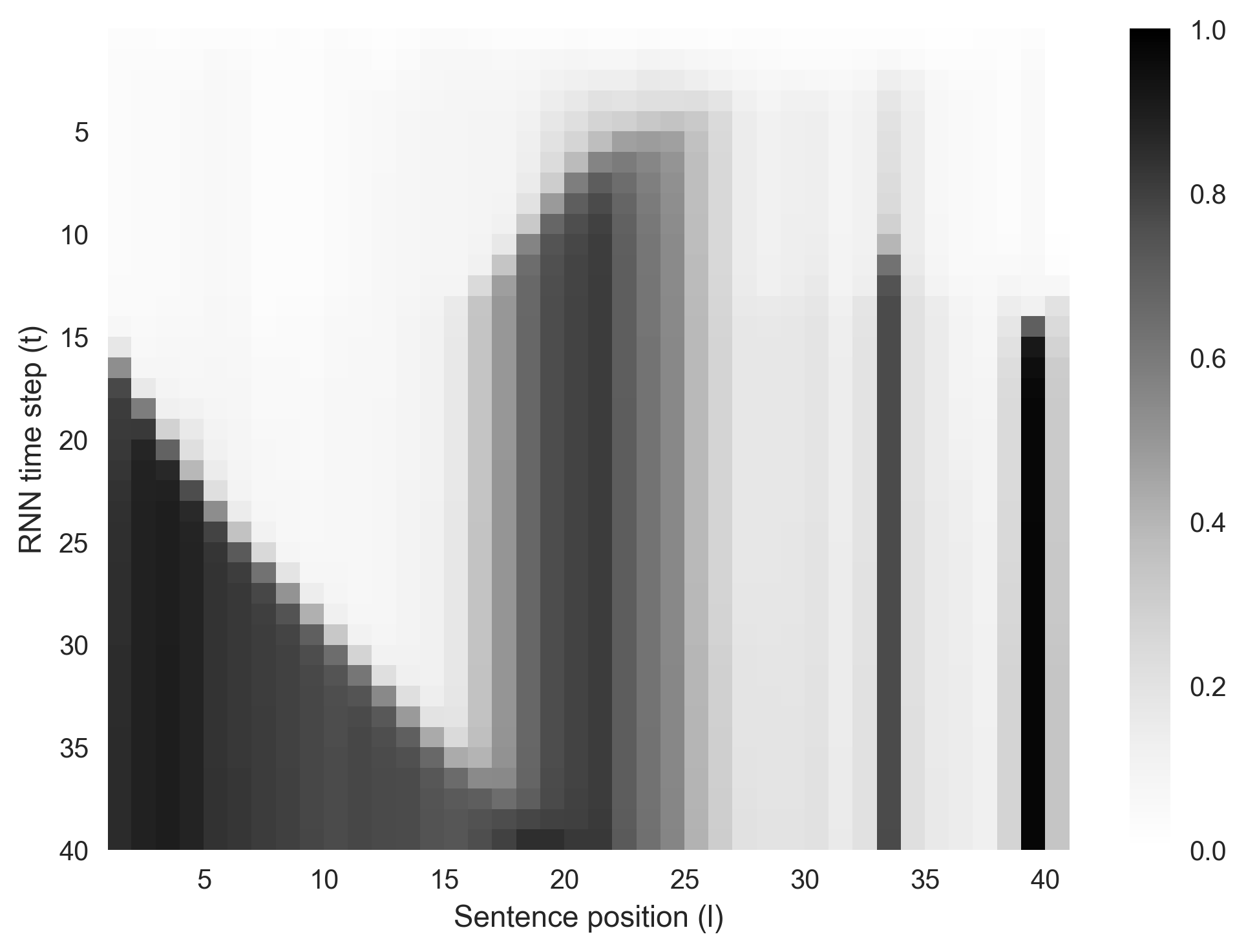}
    \caption{Visualising the cumulative attention on each of the sentence's slots for the second example of figure \ref{fig:prior_canvases}. The darker the shade, the more attention has been placed on that position.}
    \label{fig:attention}
\end{figure}

In figure \ref{fig:prior_canvases}, we show examples of how the canvas changes as the RNN makes updates to it. At each time step we take the state of the canvas and plot the sequence of words found using beam search with a beam size of 15. In figure \ref{fig:attention}, we plot the cumulative attention that has been placed on each of the canvas' slots at each RNN time step.

In figure \ref{fig:attention}, the model's attention appears to spend the first 15 time steps to decide how long the sentence will be (19 words in this case), and then spends the remaining 25 time steps filling in the words from left to right (even though it is not restricted to do so). It is notable that the model is able to dynamically adjust the length of the sentences by moving the end-of-sentence token and either inserting or deleting words as it sees fit. The model is also able to notice sentence features such as the open quotation marks at \( t = 32 \) in the second example of figure \ref{fig:prior_canvases}, which it subsequently closes at \( t = 40 \).

\subsection{Sampled sentences} \label{sec:sampled_sentences}

In tables \ref{tab:samples:prior} and \ref{tab:samples:posterior}, we show samples of text generated from the prior and posterior distributions, for both AUTR and Gen-RNN. Once again, we show the sequence of words found using beam search with a beam size of 15. In both models, sampling from the prior often produces syntactically coherent sentences. However, the AUTR samples appear to better represent the global semantics of the sentences. This is likely due to the canvas feedback mechanism when computing the RNN hidden states, and the ability of the model to place attention on the entire sentence at each time step.

When attempting to reconstruct a sentence, it appears that both models' latent representations capture information about meaning and length. For example, in the second sentence of table \ref{tab:samples:posterior}, both models are able to recognise that there is a question. The evidence of the AUTR latent representation learning meaning better than Gen-RNN is evident in several of the examples in table \ref{tab:samples:posterior}, and this is quantitatively verified by the larger contribution to AUTR's ELBO from the KL divergence term.
    
\begin{table*}[h]
    \centering
    \begin{tabular}{ l l l }
        \toprule
        \textsc{Input} \( (\mathbf{x}) \) & \textsc{Reconstruction} (AUTR) & \textsc{Reconstruction} (Gen-RNN) \\
        \toprule
        unable to stop herself, she briefly, & unable to stop herself, she leaned & unable to help her , and \\ 
        gently, touched his hand.            & forward, and touched his eyes.     & her back and her into my way. \\
        \midrule
        why didn't you tell me? & why didn't you tell me? & why didn't you tell me?'' \\ 
        \midrule
        a strange glow of sunlight shines & the light of the sun was    & a tiny light on the door,    \\ 
        down from above, paper white      & shining through the window, & and a few inches from behind \\
        and blinding, with no heat.       & illuminating the room.      & him out of the door. \\
        \midrule
        he handed her the slip of paper. & he handed her a piece of paper. & he took a sip of his drink. \\ 
        \bottomrule
    \end{tabular}
    \caption{Sentences sampled from the posterior: conditioned on a test set input sentence \( \mathbf{x} \), \( \mathbf{z} \) is drawn from its variational distribution, \( q_{\phi}(\mathbf{z}|\mathbf{x}) \), and passed through the generative model \( p_{\theta}(\mathbf{x}'|\mathbf{z}) \) to produce a reconstruction \( \mathbf{x}' \).}
    \label{tab:samples:posterior}
\end{table*}

\subsection{Imputing missing words}

AUTR's latent sentence representation makes it particularly effective at imputing missing words. To impute missing words, we use an iterative procedure inspired by the EM algorithm \cite{NH1998}. We increase a lower bound on the log-likelihood of the visible and missing data, \textit{i.e.} \( \log p_{\theta}(\mathbf{x}_{\mathrm{vis}},\mathbf{x}_{\mathrm{miss}}) \), by iterating between an E-like and M-like step, as described in algorithm \ref{alg:missing_data}. The M-like step treats the missing words as model parameters, and hence (approximately, using beam search) maximises the lower bound with respect to them.

We drop 30\% of the words from each of the test set sentences, and run algorithm \ref{alg:missing_data} with 50 different initialisations for the missing words, and select the resulting imputation with the highest bound on the log-likelihood. AUTR successfully imputes 34.1\% of the missing words, whilst Gen-RNN achieves 31.9\%. Sampled missing word imputations for AUTR and Gen-RNN are shown in table \ref{tab:samples:missing}.

\begin{algorithm}[h]
    
    \SetAlgoLined    
    
    Make an initial (random) `guess' for the missing words.
    
    \While{not converged}{%
        
        \textbf{E-like step}: Sample \( \mathbf{z} \) from its variational distribution \( q_{\phi}(\mathbf{z} | \mathbf{x}) \), where \( \mathbf{x} \) is the latest setting of the sentence.
        
        \textbf{M-like step}: Choose the missing words in \( \mathbf{x} \) to maximise \( \frac{1}{S} \sum_{s=1}^{S} \log p_{\theta}(\mathbf{x}_{\mathrm{vis}},\mathbf{x}_{\mathrm{miss}},|\mathbf{z}^{(s)}) \). This is done approximately, using beam search.
    
    }
    
    \caption{Missing data imputation}
    \label{alg:missing_data}
\end{algorithm}

\begin{table*}[h]
    \centering
    \begin{tabular}{ l l l }
        \toprule
        \textsc{Truth} & \textsc{Imputation} (AUTR) & \textsc{Imputation} (Gen-RNN) \\
        \toprule
        ``i want to draw you again,'' he says.                & ``i want to see you & ``i want to see you \\ 
        ``i want to \_\_\_\_ you \_\_\_\_\_,\_ he \_\_\_\_. & again,'' he said.   & again,'' he said.   \\
        \midrule
        i believe the lie, and so i survive another day.       & i believe the lie, and so & do believe the lie too and   \\ 
        \_ believe the lie\_ and so \_ survive another \_\_\_. & will survive another day. & so will survive another day. \\
        \midrule
        he was inside a house made of cheese.         & he was inside a house made & it is inside a house made a \\ 
        \_\_ \_\_\_ inside a house made \_\_ cheese\_ & of cheese.                 & cheese. \\
        \midrule
        i could have saved more of them if we had realized back then.              & i would have saved more & i should have saved more \\ 
        i \_\_\_\_\_ have saved more \_\_ them \_\_ we had realized back \_\_\_\_. & of them if we had       & of them than we had      \\
                                                                                   & realized back there.    & realized back then.  \\
        \bottomrule
    \end{tabular}
    \caption{Imputing missing words in test set sentences, using the procedure described in algorithm \ref{alg:missing_data}. Those words replaced with underscores (\_) are considered as missing.}
    \label{tab:samples:missing}
\end{table*}

\subsection{Exploring the latent space}

\subsubsection{Finding similar sentences}

To compare the quality of the latent representations under each model, we take a sentence \( \tilde{\mathbf{x}} \) from the test set and compute the mean of its posterior distribution, \( \pmb{\mu}_{\phi}(\tilde{\mathbf{x}}) \). We then find the `best matching' sentence \( \mathbf{x}^{*} \) in the remainder of the test set, which satisfies: \( \mathbf{x}^{*} = \arg \max_{\mathbf{x} \neq \tilde{\mathbf{x}}} \log p_{\theta}(\mathbf{x}|\mathbf{z} = \pmb{\mu}_{\phi}(\tilde{\mathbf{x}})) \). If the latent representation does indeed capture the sentence's meaning, \( \mathbf{x}^{*} \) ought to appear qualitatively similar to \( \tilde{\mathbf{x}} \).

\begin{table*}[h]
    \centering
    \begin{tabular}{@{} l l l @{}}
        \toprule
        \( \tilde{\mathbf{x}} \) & \( \mathbf{x}^{*}_{\mathrm{AUTR}} \) & \( \mathbf{x}^{*}_{\mathrm{Gen-RNN}} \) \\
        \toprule
        he wasn't ready to face the     & he was never going to see her  & she didn't want to make any      \\ 
        prospect of losing her when     & again, and that was the way it & promises, no matter how much     \\
        he'd only just gotten her back. & had to be.                     & she wanted to be with him again. \\
        \midrule
        i can't help but glare at her. & i can't help but smile at her. & i couldn't help but smile at him. \\ 
        \midrule
        so i stood in the doorway of    & as i sat on the bench outside & when he reached the bottom \\ 
        the chapel, watching it happen. & the hospital, i looked up.    & of the hill, he slowed his pace. \\
        \midrule
        dina lets a breath out on the & there is a long pause on the & he reached into his pocket and \\ 
        other side of the line.       & other end of the line.       & pulled out a piece of paper. \\
        \bottomrule
    \end{tabular}
    \caption{Finding the `best matching' sentence using the latent representation.}
    \label{tab:best_matching}
\end{table*}

We show some examples of the best matching sentences using AUTR and Gen-RNN in table \ref{tab:best_matching}; we see that the AUTR latent representations are generally successful at capturing the sentences' meanings and are able to learn sentence features such as tense and gender as well.

\subsubsection{Interpolating between latent representations}

To further understand how AUTR uses its latent representation, we randomly sample two latent representations from the prior distribution, and linearly interpolate between them. That is, we sample \( \mathbf{z}^{(1)} \) and \( \mathbf{z}^{(2)} \) from the \( \mathcal{N}(\mathbf{0}, \mathbf{I}) \) prior, and then for \( \alpha \in \{0, 0.25, 0.5, 0.75, 1\} \), we take: \begin{align}
    \mathbf{z}^{(\alpha)} = \alpha \mathbf{z}^{(1)} + (1 - \alpha) \mathbf{z}^{(2)} 
\end{align}

Then, using beam search with a beam size of 15, we evaluate the sentence produced by \( \mathbf{z}^{(\alpha)} \). Three examples are shown in table \ref{tab:latent_interpolation}; it is clear that in all of these examples, AUTR maintains the sentence's syntactic structure throughout the interpolations. It is also notable that the sentence topics and meanings remain consistent as \( \alpha \) increases. 

\begin{table*}[t!]
    \centering
    \begin{tabular}{ l | l | l | l }
        \toprule
        \( \alpha \) & Example 1 & Example 2 & Example 3 \\
        \toprule
        0 & maybe it wasn't going to happen. & ``oh, thank you.'' & one of them looked at me, \\
         & & & and smiled. \\
        \midrule
        0.25 & it would be nice to have to go. & ``oh, it's nice to meet you.'' & the moment i met her eyes, \\
         & & & she smiled at me. \\
        \midrule
        0.5 & he wasn't sure what it was about. & ``it's nice to meet you,'' he said. & the moment i stared at him, \\
         & & & he looked down at me. \\
        \midrule
        0.75 & i wasn't sure whether or not to & ``i had no idea what happened,'' & instead, he looked at me, \\
         & talk about it. & i told him. & whilst i stared at the ceiling. \\
        \midrule
        1 & i had no idea how much time to & he couldn't believe what i was & ``finally,'' he said, standing \\
         & talk about the phone calls. & talking about when i told him. & up in front of me. \\
        \bottomrule
    \end{tabular}
    \caption{Linearly interpolating between latent representations and evaluating the intermediate sentences.}
    \label{tab:latent_interpolation}
\end{table*}

\section{Conclusion} \label{sec:conclusion}

We introduce the \textit{\textbf{A}ttentive \textbf{U}nsupervised \textbf{T}ext (W)\textbf{r}iter} (AUTR), a latent variable model which uses an external memory and a dynamically updated attention mechanism to write natural language sentences. We visualise this external memory at intermediate stages to understand the sentence generation process. We find that the model achieves a higher ELBO on the Book Corpus dataset, and relies more heavily on the latent representation compared to a purely RNN-based generative model. AUTR is also computationally efficient, requiring fewer RNN time steps than the sentence length. In addition, we verify that it is able to generate coherent sentences, as well as impute missing words effectively.

We have shown that the idea of using a canvas-based mechanism to generate text is very promising and presents plenty of avenues for future research. We would like to investigate alternatives to zero-padding and the use of fixed size canvases, as well as making the number of RNN time steps dependent on the sentence's latent representation. It may also be worthwhile to consider using convolutional layers, as presented by \citeauthor{SSB2017} \shortcite{SSB2017} and \citeauthor{YHSB2017} \shortcite{YHSB2017}. A particularly interesting avenue of interest is to use an RNN with a similar attention mechanism to parametrise the variational distribution; this would also facilitate extending the model to other natural language tasks such as document classification and translation.

\section{Acknowledgements}

This work was supported by the Alan Turing Institute under the EPSRC grant EP/N510129/1.

\begin{appendices}

\section{SGVB} \label{sec:variational_inference}

Stochastic Gradient Variational Bayes (SGVB) \cite{KW2014,RMW2014} is a method for learning generative models with a joint density factorised as \( p_{\theta}(\mathbf{x}, \mathbf{z}) = p_{\theta}(\mathbf{x} | \mathbf{z}) p(\mathbf{z}) \), where \( \mathbf{x} \) is the vector of observations, \( \mathbf{z} \) is a latent vector, and \( \theta \) are the generative model parameters. The task is to learn the values of the generative parameters \( \theta \) that maximise the log-likelihood of the observed data, \textit{i.e.} \( \max_{\theta} \log p_{\theta}(\mathbf{x}) \).

For any density \( q(\mathbf{z}|\mathbf{x}) \), the evidence lower bound (ELBO) can be formed using Jensen's inequality: \begin{align}
    \log p_{\theta}(\mathbf{x}) & \geq \mathbb{E}_{q(\mathbf{z}|\mathbf{x})} \left[ \log \frac{p_{\theta}(\mathbf{z}, \mathbf{x})}{q(\mathbf{z}|\mathbf{x})} \right] \equiv \mathcal{L}(\mathbf{x}) \label{eq:elbo}
\end{align}

The optimal setting for \( q(\mathbf{z} | \mathbf{x}) \) would be the true posterior \( p(\mathbf{z}|\mathbf{x}) \), however this is usually intractable. SGVB introduces the variational parameters \( \phi \) which parametrise the distribution \( q_{\phi}(\mathbf{z} | \mathbf{x}) \). Monte Carlo sampling is used to approximate the expectation, and gradient steps are taken in the generative parameters \( \theta \) and the variational parameters \( \phi \) in order to optimise the bound.

Under certain mild conditions \cite{KW2014}, the latent vector \( \mathbf{z} \sim q_{\phi}(\mathbf{z} | \mathbf{x}) \) can be reparametrised using a differentiable transformation \( g_{\phi}(\pmb{\epsilon}, \mathbf{x}) \), for some variable \( \pmb{\epsilon} \) such that \( \mathbf{z} = g_{\phi}(\pmb{\epsilon}, \mathbf{x}) \) where \( \pmb{\epsilon} \sim p(\pmb{\epsilon}) \). The derivatives with respect to the parameters are then computed as follows, where \( \pmb{\epsilon}^{(s)} \sim p(\pmb{\epsilon}) \): \begin{align}
    \nabla_{\theta, \phi} \mathcal{L}(\mathbf{x}) &= \mathbb{E}_{p(\pmb{\epsilon})} \left[ \nabla_{\theta, \phi} \log \frac{p_{\theta}(g_{\phi}(\pmb{\epsilon}, \mathbf{x}), \mathbf{x})} {q_{\phi}(g_{\phi}(\pmb{\epsilon}, \mathbf{x}))} \right] \\
    & \simeq \frac{1}{S} \sum_{s=1}^{S} \nabla_{\theta, \phi} \log \frac{p_{\theta}(g_{\phi}(\pmb{\epsilon}^{(s)}, \mathbf{x}), \mathbf{x})} {q_{\phi}(g_{\phi}(\pmb{\epsilon}^{(s)}, \mathbf{x}))} \label{eq:approx_reparam_grad}
\end{align}

\end{appendices}

\vfill

\bibliography{refs}
\bibliographystyle{aaai}

\end{document}